\documentclass[runningheads]{llncs}
\usepackage[T1]{fontenc}
\usepackage{graphicx}
\usepackage{times}

\usepackage{soul}
\usepackage{url}
\usepackage[utf8]{inputenc}
\usepackage{booktabs}       
\usepackage{threeparttable}
\usepackage{pdflscape}
\usepackage{graphicx}
\usepackage{subfigure}
\usepackage{BOONDOX-cal}
\usepackage{mathtools}
\usepackage{newtxmath}
\usepackage{siunitx}

\usepackage{bm}
\usepackage{soul,xcolor}
\usepackage{ragged2e}
\usepackage{multirow}
\usepackage{verbatimbox}
\usepackage{tabularx} 
\usepackage{hyperref}
\usepackage[normalem]{ulem}
\usepackage{algorithm}
\usepackage{algorithmic}
\usepackage{calrsfs}
\usepackage{wrapfig}
\DeclareMathAlphabet{\pazocal}{OMS}{zplm}{m}{n}

\usepackage{amsmath,amsfonts,amssymb}
\usepackage[misc,geometry]{ifsym}
\renewcommand\footnotemark{}
\newtheorem{assumption}{Assumption}
\begin{document}
\title{Moderately-Balanced Representation Learning for Treatment Effects with Orthogonality Information}
%
\author{Yiyan Huang\inst{1} \inst{*} \and
	Cheuk Hang Leung \inst{1} \inst{*} \and
	Shumin Ma\inst{2} \and
	Qi Wu\inst{1(}\textsuperscript{\Letter}\inst{)} \and
	Dongdong Wang\inst{3} \and
	Zhixiang Huang\inst{3}
}
\thanks{\text{*} Co-first authors are in alphabetical order.}
\thanks{\text{\Letter} Qi Wu is the corresponding author.}
%
%
\institute{School of Data Science, City University of Hong Kong, Hong Kong, China \email{yiyhuang3-c@my.cityu.edu.hk, chleung87@cityu.edu.hk, qiwu55@cityu.edu.hk} \and
	Guangdong Provincial Key Laboratory of Interdisciplinary Research and Application for Data Science, BNU-HKBU United International College, Zhuhai, China\\a
	\email{shuminma@uic.edu.cn} \and
	JD Digits, Beijing, China\\
	\email{wangdongdong9@jd.com, huangzhixiang@jd.com}
}
\maketitle 
\begin{abstract}
Estimating the average treatment effect (ATE) from observational data is challenging due to selection bias. Existing works mainly tackle this challenge in two ways. Some researchers propose constructing a score function that satisfies the orthogonal condition, which guarantees that the established ATE estimator is ``orthogonal" to be more robust. The others explore representation learning models to achieve a balanced representation between the treated and the controlled groups. However, existing studies fail to 1) discriminate treated units from controlled ones in the representation space to avoid the over-balanced issue; 2) fully utilize the ``orthogonality information". In this paper, we propose a moderately-balanced representation learning (MBRL) framework based on recent covariates balanced representation learning methods and orthogonal machine learning theory. This framework protects the representation from being over-balanced via multi-task learning. Simultaneously, MBRL incorporates the noise orthogonality information in the training and validation stages to achieve a better ATE estimation. The comprehensive experiments on benchmark and simulated datasets show the superiority and robustness of our method on treatment effect estimations compared with existing state-of-the-art methods.

\keywords{Treatment effects  \and Causal inference \and Representation learning}
\end{abstract}
\section{Introduction}\label{sec:introduction}
Causal inference has drawn a lot of attention across various research areas including statistics \cite{wager2018estimation,athey2019estimating}, economics and finance \cite{farrell2015robust,chernozhukov2018double,huang2021causal} commercial social network applications \cite{guo2020ignite,DBLP:conf/kdd/ChuR021} and health care \cite{glass2013causal,hill2013assessing}. One of the main tasks of causal inference is to estimate the \textit{average treatment effect} (ATE). For example, a biotech company must know to what extent a newly developed vaccine can reduce the probability of infection for the whole population. The classical method to acquire the ATE is to conduct randomized controlled trials (RCTs), where the treatment is randomly assigned to the population but not selectively. Then the effect of the vaccine (treatment) on the infection (outcome) is measured by the difference between the average infection rate of the vaccinated group (treated group) and that of the unvaccinated group (controlled group). RCTs are regarded as the golden standard for treatment effect estimation, but conducting RCTs is costly and time-consuming \cite{pearl2009causal,guo2020survey}. Thus, estimating the treatment effects in the observational study instead of RCTs becomes more and more tempting.

When it comes to estimating the ATE from the observational data, we need to handle the selection bias. The selection bias exists due to the non-random treatment assignment. The treatment assignment may be further influenced by the covariates that also directly affect the outcome. In the vaccine example, limited vaccines tend to be distributed to vulnerable individuals who are susceptible to infection. Such a non-random treatment assignment mechanism naturally results in a covariate shift phenomenon. That is, the covariates of the treated population can substantially differ from that of the controlled population.

Two classical methods are developed for adjusting the shifted covariates: inverse propensity weighting (IPW) and regression adjustment (see more details in \cite{10.1145/3444944}). IPW weights the instances based on the propensity scores to mimic the principle of RCTs to estimate ATE. Nevertheless, the IPW estimators are sensitive to the misspecification of the propensity score. Regression adjustment methods directly estimate the outcome model instead of propensity scores, whereas they would inevitably lead to biased ATE estimations due to overfitting and regularization bias \cite{chernozhukov2018double}. Researchers improve classical methods from the perspectives of statistics and methodology. 

The \textit{orthogonal score function} proposed in \cite{chernozhukov2018double} is a statistical correction by incorporating both the outcome model and the propensity score estimations. Since such a score function satisfies the \textit{orthogonal condition}, the ATE estimator derived from the score function is consistent as long as one of the two underlying relations is correctly specified. This is also known as the \textit{doubly robust} property. Recently, balanced representation learning techniques have attracted researchers' attention. The intuitive idea is to construct a pair of ``twins" in the representation space by minimizing the imbalance between the distributions of the treated and controlled groups \cite{shalit2017estimating}. However, such methods mainly focus on the balance but overlook the discrimination between treated and controlled units. If the distributions of the treated and controlled groups in the representation space are too similar to be distinguished, it would be difficult to infer the ATE accurately. Such a trade-off plays a crucial role in identifying the treatment effects \cite{shalit2017estimating}. The importance of the undiscriminating problem is also emphasized by \cite{guo2020ignite}.

In this paper, with the tool of orthogonal machine learning, we propose a moderately-balanced representation learning (MBRL) framework to estimate the treatment effects. MBRL trains in a multi-task framework and stops on a perturbation error metric to obtain a moderately-balanced representation. The merits of MBRL include i) preserving predictive information for inferring individual outcomes; ii) designing a multi-task learning framework to achieve a moderately-balanced rather than over-balanced representation; iii) fully utilizing the orthogonality information during the training and validation stages to achieve superior treatment effect estimations.
\section{Preliminaries}
\paragraph{Potential Outcome Framework.} Let $\mathbf{Z}$ be $s$-dimensional covariates such that $\mathbf{Z} \in \pazocal{Z} \subset \mathbb{R}^s$, where $\pazocal{Z}$ is the sample space of covariates. $D \in \{0, 1\}$ denotes the treatment variable. $Y(0), Y(1)$ represent the potential outcomes for the treatment $D=0$ and $D=1$ respectively such that $Y(0), Y(1) \in \pazocal{Y} \subset \mathbb{R}$ with $\pazocal{Y}$ being the sample space of outcome. We denote $w=(\mathbf{z}, d, y)$ as the realizations of the random variables $W=(\mathbf{Z}, D, Y)$. If the observed treatment is $d$, then the factual outcome $Y^F$ equals $Y(d)$. We suppose the observational dataset contains $N$ individuals and the $m^{th}$ individual is observed as $(\mathbf{z}_m, d_m, y_m)$. The target quantity ATE $\tau$ is defined as $\tau:=\mathbb{E}\left[Y(1)-Y(0)\right]$.

Identifying the treatment effects under the potential outcome framework \cite{rubin2005causal} requires some fundamental assumptions: Strong Ignorability, Overlap, Consistency and Stable Unit Treatment Value Assumption (SUTVA). These assumptions guarantee that treatment effects can be inferred if we specify the relation $\mathbb{E}\left[Y \mid D, \mathbf{Z}\right]$, which is equivalent to estimating $g_{0}(D,\mathbf{Z})$ in the following interactive model when the treatment variable takes a binary value \cite{chernozhukov2018double}:
\begin{equation}\label{eqt:inter model}
	\begin{aligned}
		Y&=g_{0}(D,\mathbf{Z})+\xi, &&\mathbb{E}\left[\xi \mid D,\mathbf{Z}\right]=0,\\
		D&=m_{0}(\mathbf{Z}) + \nu, &&\mathbb{E}\left[\nu \mid \mathbf{Z}\right]=0.
	\end{aligned}
\end{equation}
Here, $g_0$ and $m_0$ are the true \textit{nuisance functions}. $\xi$ and $\nu$ are the noise terms. $m_{0}(\mathbf{Z})=\mathbb{E}\left[D \mid \mathbf{Z}\right]$ is the \textit{propensity score}. Let $i$ be an element of $\{0, 1\}$. The true causal parameter $\theta_{0}^i$ is defined as $\theta_{0}^i:=\mathbb{E}\left[Y(i)\right]=\mathbb{E}\left[g_0(i, \mathbf{Z})\right]$ for $i \in \{0, 1\}$, and the true ATE $\tau$ is computed by $\tau=\theta_{0}^1-\theta_{0}^0$. We denote the estimated $(\theta^{i}_{0}, g_0, m_0)$ as $(\hat{\theta}^{i}, \hat{g}, \hat{m})$, and then the estimated ATE is computed by $\hat{\tau}=\hat{\theta}^1-\hat{\theta}^0$. 

\paragraph{Orthogonal Estimators.}
We aim to estimate the true causal parameters $\theta^1_0$ and $\theta^0_0$ given $N$ i.i.d. samples $\{W_m=(\mathbf{Z}_m, D_m, Y_m)\}^{N}_{m=1}$. The standard procedure to acquire the estimated causal parameters $\hat{\theta}^1$ and $\hat{\theta}^0$ is: 1) getting the estimated nuisance functions $\hat{\rho}$, e.g., $\hat{\rho}=(\hat{g}, \hat{m})$; 2) constructing a score function $\psi(W, \theta^i, \rho)$ such that we can derive the estimated causal parameter $\hat{\theta}^i$ by solving $\mathbb{E}\left[\psi(W, \theta^i, \hat{\rho})\right]=0$, where $\theta^i$ is a causal parameter that lies in the causal parameter space. According to \cite{chernozhukov2018double}, the estimator $\hat{\theta}^i$ solved from $\mathbb{E}\left[\psi(W, \theta^i, \hat{\rho})\right]=0$ is robust to the estimated nuisance functions $\hat{\rho}$ if the corresponding score function $\psi(W, \theta^i, \rho)$ satisfies the orthogonal condition that is stated in Definition \ref{ortho_cond}.
\begin{definition}[Orthogonal Condition]\label{ortho_cond}
	Let $W=(\mathbf{Z}, D, Y)$, $\rho_0=(h_{0,1},\dots,h_{0,\gamma})$ be the true nuisance functions and $\theta_0$ be the true causal parameter with $\theta$ being a causal parameter that lies in the causal parameter space.
	A score function $\psi(W, \theta, \rho)$ is said to satisfy the orthogonal condition with respect to $\rho=(h_1,...,h_\gamma)$ if
	\begin{equation*}
		\begin{aligned}
			\mathbb{E}\left[\partial_{h_i}\psi(W, \theta, \rho)\mid_{\rho=\rho_{0}, \theta=\theta_0}  \mid \mathbf{Z} \right] = 0 \;\;\; \forall 1 \leq i \leq \gamma.
		\end{aligned}
	\end{equation*}
\end{definition}

Under the interactive model setup \eqref{eqt:inter model}, the nuisance functions are $(g, m)$, and the true ones are $(g_0, m_0)$.
In this case, the orthogonal condition guarantees that the estimator is consistent if either one of the two nuisance functions, but unnecessarily both, is accurately estimated. This is well known as the doubly robust property. In this paper, we introduce two orthogonal estimators $\hat{\theta}_{1}$ \cite{chernozhukov2018double} and $\hat{\theta}_{2}$ \cite{huang2021higher} in Proposition \ref{score function}, and we can estimate ATE by plugging the learned nuisance functions into the orthogonal estimators.

\begin{proposition}[Orthogonal Estimators]\label{score function}
Let the nuisance functions be $\rho=(g,m)$ and the causal parameter be $\theta^i$ for $i \in \{0, 1\}$, the score functions $\psi_{1}(W, \theta^{i}, \rho)$ and $\psi_{2}(W, \theta^{i}, \rho)$ that satisfy the orthogonal condition (Definition \ref{ortho_cond}) are:
\begin{align}
		&\psi_{1}(W, \theta^{i}, \rho) = \theta^{i}-g(i,\mathbf{Z}) \label{eqt:ortho_score func_1}-(Y-g(i,\mathbf{Z}))\frac{iD+(1-i)(1-D)}{im(\mathbf{Z})+(1-i)(1-m(\mathbf{Z}))}; \\
		&\psi_{2}(W, \theta^{i}, \rho) = \theta^{i}-g(i,\mathbf{Z}) \label{eqt:ortho_score func_2}-(Y(i)-g(i,\mathbf{Z}))\frac{\left((D-m(\mathbf{Z}))-\mathbb{E}\left[\nu \mid \mathbf{Z} \right]\right)^2}{\mathbb{E}\left[\nu^2 \mid \mathbf{Z}\right]}. 
	\end{align}
The corresponding orthogonal estimators are:
\begin{align}
		\hat{\theta}_1^i \quad \text{solves} \quad \frac{1}{N}\sum\limits_{m=1}^{N}\psi_1(W_m, \theta^{i}, \hat{\rho})=0; \quad
		\hat{\theta}_2^i \quad \text{solves} \quad \frac{1}{N}\sum\limits_{m=1}^{N}\psi_2(W_m, \theta^{i}, \hat{\rho})=0. \nonumber
\end{align}
\end{proposition}


\section{Method}\label{sec:method}
In this section, we first introduce the orthogonality information in Section \ref{sec:theoretical results}. Then we present the network structure, objective function and model selection criterion of the proposed MBRL method based on the orthogonality information in Section \ref{sec:the proposed framework}.

\subsection{Orthogonality Information} \label{sec:theoretical results}
Recall that the ATE estimators $\hat{\theta}^i_{1}$ and $\hat{\theta}^i_{2}$ are doubly robust since they are orthogonal estimators. Still, they could be non-orthogonal once the model setup \eqref{eqt:inter model} relaxes the restrictions on the noise terms $\xi$ and $\nu$ since the score functions $\psi_1$ and $\psi_2$ might violate the orthogonal condition. Hence, we propose the \textit{Noise Conditions}, which would enforce the learned nuisance functions adapted to orthogonal estimators.
\begin{proposition}[Noise Conditions]\label{noise conditions}
	Under the interactive model setup \eqref{eqt:inter model}, the conditions on the noise terms $\xi$ and $\nu$, i.e., $\mathbb{E}\left[\xi \mid D, \mathbf{Z} \right]=0$ and $\mathbb{E}\left[\nu \mid \mathbf{Z} \right]=0$, are sufficient conditions for $\psi_1$ and $\psi_2$ being orthogonal score functions ($\hat{\theta}_{1}^i$ and $\hat{\theta}_{2}^i$ being orthogonal estimators).
\end{proposition}
Given the noise conditions, we can exploit an essential property, the \textit{noise orthogonality} property.
\begin{property}[Noise Orthogonality]\label{noise orthogonality}
	Under the interactive model setup \eqref{eqt:inter model} and the noise conditions, we have $\mathbb{E}[(Y-g_0(D, \mathbf{Z}))(D-m_0(\mathbf{Z}))]=0$.
\end{property}
The noise conditions are sufficient conditions for the estimators $\hat{\theta}_{1}^i$ and $\hat{\theta}_{2}^i$ being orthogonal, so noise conditions play an important role when we approximate the true nuisance functions $(g_0, m_0)$ with estimated ones $(\hat{g}, \hat{m})$. Besides, under the noise conditions, the noise orthogonality can be utilized for our model selection. The decompositions similar to Noise Orthogonality also appeared in \cite{chernozhukov2018double,10.1145/3459637.3482339}.
\subsection{The Proposed Framework}\label{sec:the proposed framework}
\begin{figure}[h]
	\centering	
	\includegraphics[width=0.8\columnwidth]{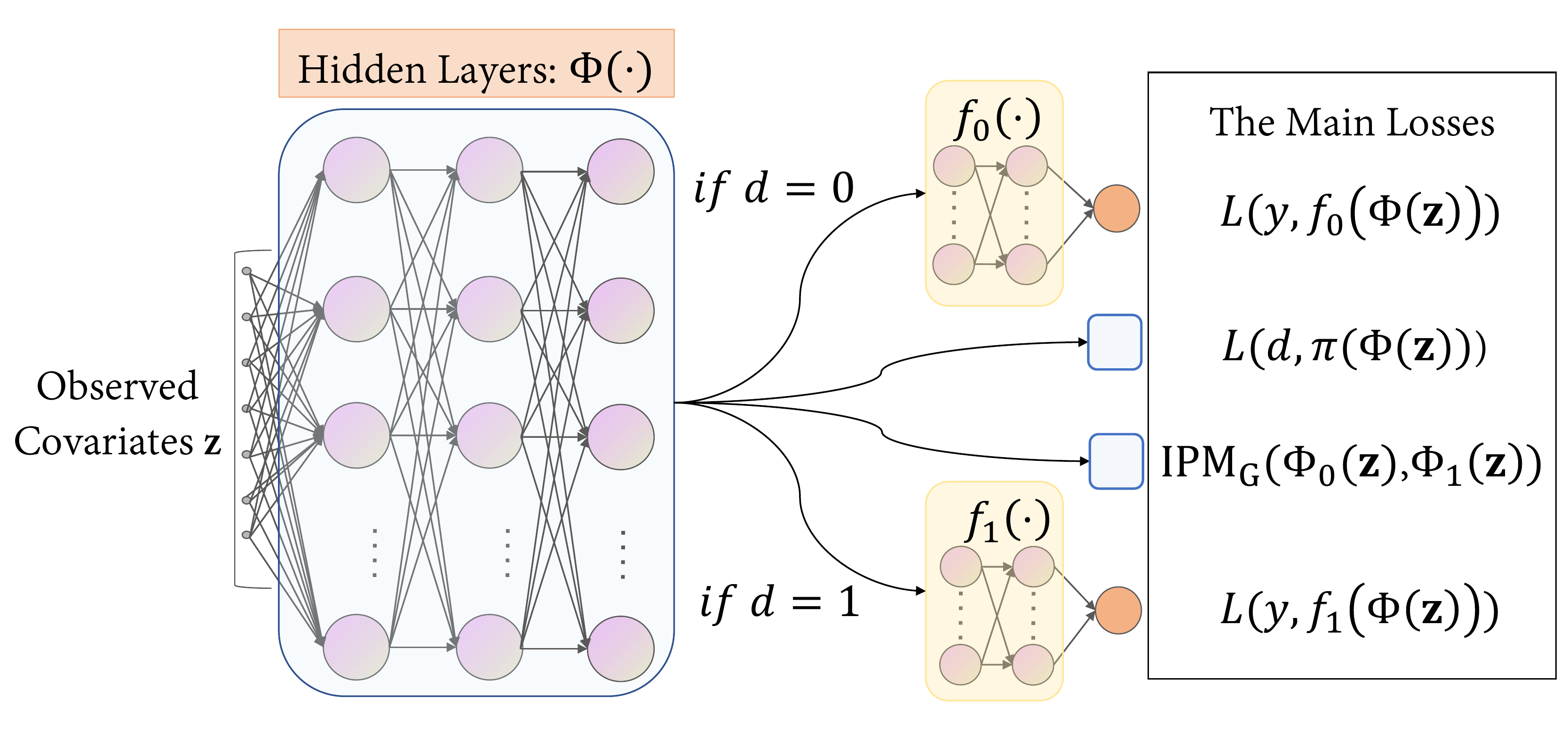}
	\caption{The MBRL network architecture.}
	\label{Figure:model}
\end{figure}

We propose a moderately-balanced representation learning (MBRL) framework to obtain $(\hat{g}, \hat{m})$ to estimate ATE, and the MBRL architecture is illustrated in Figure \ref{Figure:model}. The MBRL network maps the original covariates space to the representation space (i.e., $\Phi:\pazocal{Z} \rightarrow \pazocal{R}$) such that 1) the representation preserves predictive information for outcomes; 2) the map makes the distributional discrepancy between the treated group and the controlled group small enough; 3) the domain (treated or controlled) of each individual is well discriminated; 4) the orthogonality information is involved.
\vspace{-0.5cm}
\vspace{-0.5cm}
\paragraph{Learning Representation of Covariates.}
The distributions of the treated group and the controlled group are inherently disparate due to selection bias. Previous works handle this problem using a balanced representation learning method \cite{johansson2016learning,shalit2017estimating}, which forces the distributions of treatment and control groups to be similar enough in the representation space. Specifically, a representation is learned by minimizing the integral probability metrics (IPM), which measures the imbalance between the distributions of the treated population and the controlled population (see the details in \cite{shalit2017estimating}):
\begin{equation}
	\begin{aligned}
		&\pazocal{L}_{imb} = \text{IPM}_\pazocal{G}(\{\Phi(\mathbf{z}_m)\}_{m:d_m=1},\{\Phi(\mathbf{z}_m)\}_{m:d_m=0}).
	\end{aligned}
\end{equation}
\paragraph{The Prediction of Outcome and Treatment.} MBRL predicts the outcome by the function $f: \{0,1\} \times \pazocal{R} \rightarrow \pazocal{Y}$, which is partitioned into two functions $f_0$ and $f_1$:
\begin{equation}
{\small\begin{aligned}
	f(d_m, \Phi(\mathbf{z}_m))=d_mf_1(\Phi(\mathbf{z}_m))+(1-d_m)f_0(\Phi(\mathbf{z}_m)).
\end{aligned}}	
\end{equation}
$f_1$ and $f_0$ are the output functions that map the representation to the potential outcomes for $D=1$ and $D=0$, respectively. $f(d_m, \Phi(\mathbf{z}_m))$ is the predicted factual outcome and we aim to minimize the factual outcome loss $\pazocal{L}_{fo}$ such that
\begin{equation}
{\small\begin{aligned}\label{eqt:L_FO}
		\pazocal{L}_{fo} = \frac{1}{N}\sum_{m=1}^{N}\left[y_m-f(d_m, \Phi(\mathbf{z}_m))\right]^2. 
	\end{aligned}}
\end{equation}
Here, $\hat{g}(d_m, \mathbf{z}_m)=f(d_m, \Phi(\mathbf{z}_m))$ is the estimated factual outcome of the $m^{th}$ unit. Aside from making a low-error prediction over factual outcomes with a small divergence between treated and controlled groups, the distinguishability of the treated units from the controlled ones is also non-negligible. Therefore, we propose to maximize the distinguishability loss $\pazocal{L}_{dis}$ (measured by log-likelihood) such that
\begin{equation}\label{eqt:L_FT}
{\small	\begin{aligned}
		\pazocal{L}_{dis}=\frac{1}{N}&\sum_{m=1}^{N}\big[d_m\log\pi(\Phi(\mathbf{z}_m))
		 + (1-d_m)\log(1-\pi(\Phi(\mathbf{z}_m)))\big].
	\end{aligned}}
\end{equation}
Here, $\hat{m}(\mathbf{z}_m)=\pi(\Phi(\mathbf{z}_m))$ is the estimated probability of the $m^{th}$ unit being assigned the treatment $D=1$ (aka the estimated propensity score).

\paragraph{The Noise Regularizations.}
Recall Proposition \ref{noise conditions} that $\mathbb{E}\left[\xi \mid D, \mathbf{Z}\right]=0$ and $\mathbb{E}\left[\nu \mid \mathbf{Z}\right]=0$ are sufficient conditions for score functions $\psi_{1}$ and $\psi_{2}$ being orthogonal. Empirically, we want to involve the following constraints:
\begin{equation}\label{noise constraint}
{\small	\begin{aligned}
		& \frac{1}{N}\sum_{m=1}^{N}\left[y_m-f(d_m, \Phi(\mathbf{z}_m))\right]=0,\\ 
		& \frac{1}{N}\sum_{m=1}^{N}\left[d_m-\pi(\Phi(\mathbf{z}_m))\right]=0.
	\end{aligned}}
\end{equation}
This motivates us to formalize $\Omega_{y}$ and $\Omega_{d}$ such that
\begin{equation}\label{eqt:NOR}
{\small	\begin{aligned}
		& \Omega_{y} = \epsilon_y \big| \frac{1}{N}\sum_{m=1}^{N}[y_m-f(d_m, \Phi(\mathbf{z}_m))] \big|, \\
		& \Omega_{d} = \epsilon_d \big| \frac{1}{N}\sum_{m=1}^{N}[d_m-\pi(\Phi(\mathbf{z}_m))] \big|.
	\end{aligned}}
\end{equation}
The partial derivative of $\Omega_{y}$ w.r.t. $\epsilon_y$ (or $\Omega_{d}$ w.r.t. $\epsilon_d$) equaling $0$ forces the learned nuisance functions to satisfy Eqn. \eqref{noise constraint}. Therefore, minimizing the noise regularizations $\Omega_{y}$ and $\Omega_{d}$ adapts the entire learning process to satisfy the orthogonal score function. This idea corresponds to the targeted regularizations (see more discussions in \cite{shi2019adapting,10.1145/3459637.3482339}).

\paragraph{Multi-task Learning and Perturbation Error.}
MBRL learns the nuisance functions through multi-task learning with following three tasks in each iteration:
\begin{equation}\label{eqt:final loss}
	\begin{aligned}
		&\text{Task 1:}	\;\;\;\;\;\;\;\;\;	\max_{\pi,\epsilon_d} \;\;\; \pazocal{L}_{dis} - \lambda_1\Omega_{d}\\
		&\text{Task 2:}	\;\;\;\;\;\;\;\;\;	\	\min_{\Phi} \;\;\; \pazocal{L}_{imb}\\
		&\text{Task 3:}	\;\;\;\;\;\;\;\;\;		\min_{\Phi,f,\epsilon_y} \;\;\; \pazocal{L}_{fo} + \lambda_2\Omega_{y}\\
	\end{aligned}
\end{equation}
Instead of putting $\pazocal{L}_{imb}$ into Task 3 as a regularization, we let $\pazocal{L}_{imb}$ be one of the multiple tasks. To be specific, Task 1 updates $\pi$ to produce the propensity scores, and Task 2 achieves a balance between $\{\Phi(\mathbf{z}_m)\}_{m:d_m=1}$ and $\{\Phi(\mathbf{z}_m)\}_{m:d_m=0}$. Additionally, MBRL incorporates a novel model selection criterion, the \textit{Perturbation Error}, according to the noise orthogonality property. It takes advantage of the noise orthogonality information by perturbating the main evaluation metric. For example, if the final model is selected by the metric root-mean-square error ({\small$RMSE = \sqrt{\frac{1}{N}\sum_{m=1}^{N}(y_m-\hat{y}_m)^2}$}), then the perturbation error $\epsilon_{p}$ is defined as
{\small	\begin{align}
		\epsilon_{p}=RMSE + \beta|\frac{1}{N}\sum_{m=1}^{N}(y_m-\hat{y}_m)(d_m-\hat{d}_m)|. \nonumber
\end{align}}
Here, $\beta$ is the perturbation coefficient which is a constant; $\hat{y}_m$ and $\hat{d}_m$ are the predicted values of $f(d_m, \Phi(\mathbf{z}_m))$ and $\pi(\Phi(\mathbf{z}_m))$, respectively. The final model is selected on the validation set based on the minimum $\epsilon_{p}$. If either outcome or propensity score is well specified (i.e., representations are moderately-balanced instead of over-balanced), the second term in $\epsilon_{p}$ would be small.

\section{Experiments}\label{sec:experiments}
In this section, we conduct comprehensive experiments on benchmark datasets to evaluate the performance produced by MBRL and other prevalent causal inference methods. We further test the effectiveness of MBRL on simulated datasets with different levels of selection bias. All the experiments are run on Dell 7920 with 1x 16-core Intel Xeon Gold 6250 3.90GHz CPU and 3x NVIDIA Quadro RTX 6000 GPU.
\subsection{Dataset Description}
Since the ground truth of treatment effects are inaccessible for real-world data, it is difficult to evaluate the performance of causal inference methods for ATE estimation. Previous causal inference literatures assess their methods on two prevalent semi-synthetic datasets: IHDP and Twins.
\paragraph{IHDP.} The IHDP dataset is a well-known benchmark dataset for causal inference introduced by \cite{hill2011bayesian}. It includes 747 samples with 25-dimensional covariates associated with the information of infants and their mothers, such as birth weight and mother's age. These covariates are collected from a real-world randomized experiment. Our aim is to study the treatment effect of the specialist visits (binary treatment) on the cognitive scores (continuous-valued outcome). The outcome is generated using the NPCI package \cite{npci}, and the selection bias is created by removing a subset of the treated population. We use the same 1000 IHDP datasets as the ones used in \cite{shalit2017estimating}, where each dataset is split by the ratio of $63\%/27\%/10\%$ as training/validation/test sets.

\paragraph{Twins.}
The Twins dataset \cite{louizos2017causal} collects twin births in the USA between 1989 and 1991 \cite{almond2005costs}. After the data processing, each unit has 30 covariates relevant to parents, pregnancy and birth \cite{yoon2018ganite}. The treatment $D=1$ indicates the heavier twin while $D=0$ indicates the lighter twin, and the outcome $Y$ is a binary variable defined as the 1-year mortality. Similar to \cite{yoon2018ganite}, we only select twins who have the same gender and both weigh less than 2 kg, which finally gives $11440$ pairs of twins whose mortality rate is $17.7\%$ for the lighter twin, and $16.1\%$ for the heavier twin. To create the selection bias, we selectively choose one of the two twins as the factual observation based on the covariates of $m^{th}$ individual: $D_m|\mathbf{Z}_m \sim$ Bernoulli(Sigmoid($\mathbf{w}^T\mathbf{Z}_m+n$)), where $\mathbf{w} \sim \pazocal{U}((-0.01,0.01)^{30\times1})$ and $n \sim \pazocal{N}(0,0.01)$. We repeat the data generating process for $100$ times, and the generated $100$ Twins datasets are all split by the ratio of $56\%/24\%/20\%$ as training/validation/test sets.

\subsection{Performance Measurement and Experimental settings}
\paragraph{Performance Measurement.}
Generally, the comparisons are based on the absolute error in ATE: $\epsilon_{ATE}=|\tau-\hat{\tau}|$. Additionally, we also test the performance of MBRL on individual treatment effect (ITE) estimations. For IHDP datasets, we adopt Precision in Estimation of Heterogeneous Effect (PEHE): $$\epsilon_{PEHE}=\frac{1}{N}\sum_{m=1}^{N}\left([y_m(1)-y_m(0)]-[\hat{y}_m(1)-\hat{y}_m(0)]\right)^2.$$ For Twins datasets, we follow \cite{louizos2017causal} to adopt Area Under ROC Curve (AUC).

\paragraph{Baseline Models.} We compare our MBRL method with the following basline models: linear regression with the treatment as feature (\textbf{OLS/LR$_1$}), separate linear regression for each treatment group (\textbf{OLS/LR$_2$}), k-nearest neighbor (\textbf{k-NN}), bayesian additive regression trees (\textbf{BART}) \cite{chipman2010bart}, causal forest (\textbf{CF}) \cite{wager2018estimation}, balancing linear regression (\textbf{BLR}) \cite{johansson2016learning}, balancing neural network (\textbf{BNN}) \cite{johansson2016learning}, treatment-agnostic representation network (\textbf{TARNet}) \cite{shalit2017estimating}, counterfactual regression with Wasserstein distance (\textbf{CFR-WASS}) \cite{shalit2017estimating}, causal effect
variational autoencoders (\textbf{CEVAE}) \cite{louizos2017causal}, local similarity preserved individual treatment effect (\textbf{SITE}) \cite{yao2018representation}, generative adversarial networks for inference of treatment effect (\textbf{GANITE}) \cite{yoon2018ganite} and (\textbf{Dragonnet}) \cite{shi2019adapting}.
\paragraph{Experimental Details.}
In our experiments, $\text{IPM}_{\pazocal{G}}$ is chosen as the Wasserstein distance. Let the empirical distribution of representation be $P(\Phi(\mathbf{Z}))=P(\Phi(\mathbf{Z}) \mid D=1)$ for the treated group and $Q(\Phi(\mathbf{Z}))=Q(\Phi(\mathbf{Z}) \mid D=0)$ for the controlled group. Assuming that $\pazocal{G}$ is defined as the functional space of a family of 1-Lipschitz functions, we obtain the 1-Wasserstein distance for $\text{IPM}_{\pazocal{G}}$ \cite{shalit2017estimating}:
\begin{equation*}
	\begin{aligned}
		&Wass(P,Q)
		=\inf_{k \in \pazocal{K}} \int_{\mathbf{h} \in \{\Phi(\mathbf{Z}_m)\}_{m:D_m=1}} \|k(\mathbf{h})-\mathbf{h}\| P(\mathbf{h}) d\mathbf{h}.
	\end{aligned}
\end{equation*}
Here, $\pazocal{K}=\{k \mid k: Q(k(\Phi(\mathbf{Z})))=P(\Phi(\mathbf{Z}))\}$ defines the set of push-forward functions that transform the representation
distribution of the treated group $P(\Phi(\mathbf{Z}))$ to that of the controlled group $Q(\Phi(\mathbf{Z}))$.
\begin{table}[h]
	\centering
	\caption{Performance comparisons and ablation study with mean $\pm$ standard error on 1000 \textbf{IHDP} datasets. $\epsilon_{ATE}$: Lower is better. $\sqrt{\epsilon_{PEHE}}$: Lower is better.}
	\resizebox{0.7\columnwidth}{!}{
		\begin{tabular}{ccccc}
			\toprule
			\multirow{2}[2]{*}{Method} & \multicolumn{2}{c}{In-sample} & \multicolumn{2}{c}{Out-of-sample} \\
			& $\sqrt{\epsilon_{PEHE}}$ & $\epsilon_{ATE}$ & $\sqrt{\epsilon_{PEHE}}$ & $\epsilon_{ATE}$ \\
			\midrule
			OLS/LR$_1$ & $5.8 \pm .3$ & $.73 \pm .04$  & $5.8 \pm .3$ & $.94 \pm .06$ \\
			OLS/LR$_2$ & $2.4 \pm .1$ & $.14 \pm .01$ & $2.5 \pm .1$ & $.31 \pm .02$ \\
			k-NN  & $2.1 \pm .1$ & $.14 \pm .01$ & $4.1 \pm .2$ & $.79 \pm .05$ \\
			BART  & $2.1 \pm .1$ & $.23 \pm .01$ & $2.3 \pm .1$ & $.34 \pm .02$ \\
			CF    & $3.8 \pm .2$ & $.18 \pm .01$ & $3.8 \pm .2$ & $.40 \pm .03$ \\
			CEVAE & $2.7 \pm .1$ & $.34 \pm .01$ & $2.6 \pm .1$ & $.46 \pm .02$ \\
			SITE  & $.69 \pm .0$ & $.22 \pm .01$ & $.75 \pm .0$ & $.24 \pm .01$ \\
			GANITE & $1.9 \pm .4$ & $.43 \pm .05$ & $2.4 \pm .4$ & $.49 \pm .05$ \\
			BLR   & $5.8 \pm .3$ & $.72 \pm .04$ & $5.8 \pm .3$ & $.93 \pm .05$ \\
			BNN   & $2.2 \pm .1$ & $.37 \pm .03$ & $2.1 \pm .1$ & $.42 \pm .03$ \\
			TARNet & $.88 \pm .0$ & $.26 \pm .01$ & $.95 \pm .0$ & $.28 \pm .01$ \\
			CFR-WASS & $.71 \pm .0$ & $.25 \pm .01$ & $.76 \pm .0$ & $.27 \pm .01$ \\
			Dragonnet & $1.3 \pm .4$ & $.14 \pm .01$ & $ 1.3 \pm .5$ & $.20 \pm .05$ \\
			\midrule
			MBRL  & \boldmath{}\textbf{$.52 \pm .0$}\unboldmath{} & $.12 \pm .01$ & \boldmath{}\textbf{$.57 \pm .0$}\unboldmath{} & \boldmath{}\textbf{$.13 \pm .01$}\unboldmath{} \\
			MBRL + $\hat{\theta}^i_{1}$ & \boldmath{}\textbf{$.52 \pm .0$}\unboldmath{} & \boldmath{}\textbf{$.10 \pm .00$}\unboldmath{} & \boldmath{}\textbf{$.57 \pm .0$}\unboldmath{} & $.17 \pm .01$ \\
			MBRL + $\hat{\theta}^i_{2}$ & \boldmath{}\textbf{$.52 \pm .0$}\unboldmath{} & $.11 \pm .00$ & \boldmath{}\textbf{$.57 \pm .0$}\unboldmath{} & $.20 \pm .01$ \\
			\bottomrule
		\end{tabular}%
	}
	\label{tab:IHDP}%
\end{table}%
In addition, we adopt ELU activation function and set 4 fully connected layers with 200 units for both the representation encoder network $\Phi(\cdot)$ and the discriminator $\pi(\cdot)$, and 3 fully connected layers with 100 units for the outcome prediction networks $f_0(\cdot)$ and $f_1(\cdot)$. The optimizer is chosen as Adam \cite{kingma2014adam}, and the learning rate for the optimizer is set to be $1e^{-3}$. We set (batch size, epoch) to be $(100, 1000)/(1000, 250)$ for IHDP/Twins experiments, and the hyper parameters $(\lambda_1,\lambda_2)$ to be $(0.01, 0.01)/(0.1,0.1)$ for IHDP/Twins experiments. The final model early stops on the metric $\epsilon_p$, and we choose $\beta$ in $\epsilon_p$ as $0.1$ and $100$ for IHDP experiments and Twins experiments, respectively.

For the baseline models, we follow the same settings of hyperparameters as in their published paper and code. For our MBRL network, the optimal hyperparameters are chosen in the same way as \cite{shalit2017estimating}. The searching ranges are reported in Table \ref{table:hyper}.

\begin{table}[h]
	\centering
	\caption{Performance comparisons with mean $\pm$ standard error on 100 \textbf{Twins} datasets. $\epsilon_{ATE}$: Lower is better. AUC: Higher is better.}
	\resizebox{0.7\columnwidth}{!}{
		\begin{tabular}{ccccc}
			\toprule
			\multirow{2}[2]{*}{Method} & \multicolumn{2}{c}{In-sample} & \multicolumn{2}{c}{Out-of-sample} \\
			& AUC   & $\epsilon_{ATE}$ & AUC   & $\epsilon_{ATE}$ \\
			\midrule
			OLS/LR$_1$ & $.660 \pm .005$ & $.004 \pm .003$ & $.500 \pm .028$ & $.007 \pm .006$ \\
			OLS/LR$_2$ & $.660 \pm .004$ & $.004 \pm .003$ & $.500 \pm .016$ & $.007 \pm .006$ \\
			k-NN  & $.609 \pm .010$ & \boldmath{}\textbf{$.003 \pm .002$}\unboldmath{} & $.492 \pm .012$ & \boldmath{}\textbf{$.005 \pm .004$}\unboldmath{} \\
			BART  & $.506 \pm .014$ & $.121 \pm .024$ & $.500 \pm .011$ & $.127 \pm .024$ \\
			CEVAE & $.845 \pm .003$ & $.022 \pm .002$ & $.841 \pm .004$ & $.032 \pm .003$ \\
			SITE  & $.862 \pm .002$ & $.016 \pm .001$ & $.853 \pm .006$ & $.020 \pm .002$ \\
			BLR   & $.611 \pm .009$ & $.006 \pm .004$ & $.510 \pm .018$ & $.033 \pm .009$ \\
			BNN   & $.690 \pm .008$ & $.006 \pm .003$ & $.676 \pm .008$ & $.020 \pm .007$ \\
			TARNet & $.849 \pm .002$ & $.011 \pm .002$ & $.840 \pm .006$ & $.015 \pm .002$ \\
			CFR-WASS & $.850 \pm .002$ & $.011 \pm .002$ & $.842 \pm .005$ & $.028 \pm .003$ \\
			\midrule
			MBRL  & \boldmath{}\textbf{$.879 \pm .000$}\unboldmath{} & \boldmath{}\textbf{$.003 \pm .000$}\unboldmath{} & \boldmath{}\textbf{$.874 \pm .001$}\unboldmath{} & $.007 \pm .001$ \\
			MBRL + $\hat{\theta}^i_{1}$ & \boldmath{}\textbf{$.879 \pm .000$}\unboldmath{} & \boldmath{}\textbf{$.003 \pm .000$}\unboldmath{} & \boldmath{}\textbf{$.874 \pm .001$}\unboldmath{} & $.008 \pm .000$ \\
			MBRL + $\hat{\theta}^i_{2}$ & \boldmath{}\textbf{$.879 \pm .000$}\unboldmath{} & \boldmath{}\textbf{$.003 \pm .000$}\unboldmath{} & \boldmath{}\textbf{$.874 \pm .001$}\unboldmath{} & $.006 \pm .001$ \\
			\bottomrule
		\end{tabular}%
	}
	\label{tab:Twins}%
\end{table}%
\subsection{Results Analysis} Table \ref{tab:IHDP} and Table \ref{tab:Twins} report part of the performances of baseline methods and MBRL on IHDP and Twins datasets. We present the average values and standard errors of $\epsilon_{ATE}$, $\epsilon_{PEHE}$ and AUC (mean $\pm$ std). The lower $\epsilon_{ATE}$ and $\epsilon_{PEHE}$ or the higher AUC, the better. Bold indicates the best method for each dataset.

As stated in Table \ref{tab:IHDP} and Table \ref{tab:Twins}, we have the following observations. 1) MBRL achieves significant improvements in both ITE and ATE estimations across all datasets compared to the baseline models. 2) The advanced representation learning methods that focus on estimating ITE (such as SITE, TARNet and CFR-WASS) show their inapplicability to ATE estimations. By contrast, MBRL not only significantly outperforms these representation learning methods in ITE estimations but also remains among the best ATE results. 3) The state-of-the-art ATE estimation method, Dragonnet, achieves superior ATE estimations across all the baseline models but yields a substantial error in ITE estimations. Although Dragonnet shares a similar basic network architecture to MBRL, MBRL can obtain a substantially lower $\epsilon_{ATE}$ than Dragonnet owing to the multi-task learning framework and the utilization for orthogonality information. These observations indicate that the proposed MBRL method is extremely effective for estimating treatment effects.

We further conduct an ablation study on IHDP datasets to test if orthogonality information is practical in real applications. The relevant results are reported in Table \ref{tab:ablation}. We let MBRL* denote MBRL without perturbation error $\epsilon_{p}$, and MBRL** denote MBRL without any orthogonality information ($\epsilon_{p}$, $\Omega_{d}$ and $\Omega_{y}$). We find that incorporating orthogonality information will enhance the power of estimating treatment effects, whether with or without orthogonal estimators. This enhancement is pronounced especially when orthogonal estimators are plugged in for in-sample data.
\begin{table}[htbp]
	\begin{minipage}{0.57\columnwidth}
	\centering
	\caption{Ablation study on IHDP datasets.}
	\resizebox{1\columnwidth}{!}{
	\begin{tabular}{ccccc}
		\toprule
		\multirow{2}[2]{*}{Method} & \multicolumn{2}{c}{In-sample} & \multicolumn{2}{c}{Out-of-sample} \\
		& $\sqrt{\epsilon_{PEHE}}$ & $\epsilon_{ATE}$ & $\sqrt{\epsilon_{PEHE}}$ & $\epsilon_{ATE}$ \\
		\midrule
		MBRL** & $.523 \pm .006$ & $.129 \pm .005$ & $.568 \pm .009$ & $.141 \pm .006$ \\
		MBRL* & $.522 \pm .006$ & $.128 \pm .005$ & $.567 \pm .009$ & $.139 \pm .006$ \\
		MBRL  & $.522 \pm .007$ & $.121 \pm .005$ & $.565 \pm .008$ & $.133 \pm .005$ \\
		\midrule
		MBRL** + $\theta^i_{1}$ & $.523 \pm .006$ & $.101 \pm .004$ & $.568 \pm .009$ & $.171 \pm .007$ \\
		MBRL* + $\theta^i_{1}$ & $.523 \pm .006$ & $.102 \pm .004$ & $.567 \pm .009$ & $.170 \pm .007$ \\
		MBRL + $\theta^i_{1}$ & $.522 \pm .007$ & $.102 \pm .004$ & $.565 \pm .008$ & $.166 \pm .007$ \\
		\midrule
		MBRL** + $\theta^i_{2}$ & $.523 \pm .006$ & $.122 \pm .005$ & $.568 \pm .009$ & $.210 \pm .008$ \\
		MBRL* + $\theta^i_{2}$ & $.523 \pm .006$ & $.121 \pm .005$ & $.567 \pm .009$ & $.208 \pm .008$ \\
		MBRL + $\theta^i_{2}$ & $.522 \pm .007$ & $.114 \pm .005$ & $.565 \pm .008$ & $.204 \pm .008$ \\
		\bottomrule
	\end{tabular}%
}
	\label{tab:ablation}%
\end{minipage}
	\begin{minipage}{0.41\columnwidth}
		\centering
		\caption{The searching ranges of hyperparameters.}
		\resizebox{1\columnwidth}{!}{
				\begin{tabular}{lcc}
						\toprule
						\multicolumn{1}{c} Hyperparameters & IHDP  & Twins \\
						\midrule
						$\lambda_1, \lambda_2$ & $0.01, 0.1, 1$ & $0.01, 0.1, 1$ \\
						Depth of $\Phi$ & $2, 3, 4$ & $2, 3, 4$ \\
						Dim of $\Phi$ & $100, 200$ & $100, 200$ \\
						Depth of $\pi$ & $2, 3, 4$ & $2, 3, 4$ \\
						Dim of $\pi$ & $100, 200$ & $100, 200$ \\
						Depth of $f_0$, $f_1$ & $2, 3, 4$ & $2, 3, 4$ \\
						Dim of $f_0$, $f_1$ & $100, 200$ & $100, 200$ \\
						Batch size & $100, 300$ & $500, 1000$ \\
						Epoch & $500, 1000$ & $250, 500$ \\
						\bottomrule
					\end{tabular}%
			}
		\label{table:hyper}%
	\end{minipage}%
\end{table}%
\subsection{Simulation Study} In this part, we mainly investigate two questions. Q1. Does MBRL perform more stably to the level of selection bias than the state-of-the-art model Dragonnet? Q2. Can the noise orthogonality information, the perturbation error $\epsilon_{p}$, improve ATE estimations regardless of different models/estimators/selection bias levels?

\begin{figure}[htbp]
	\centering
	\subfigure[MBRL vs. Dragonnet]{\includegraphics[width=0.49\columnwidth]{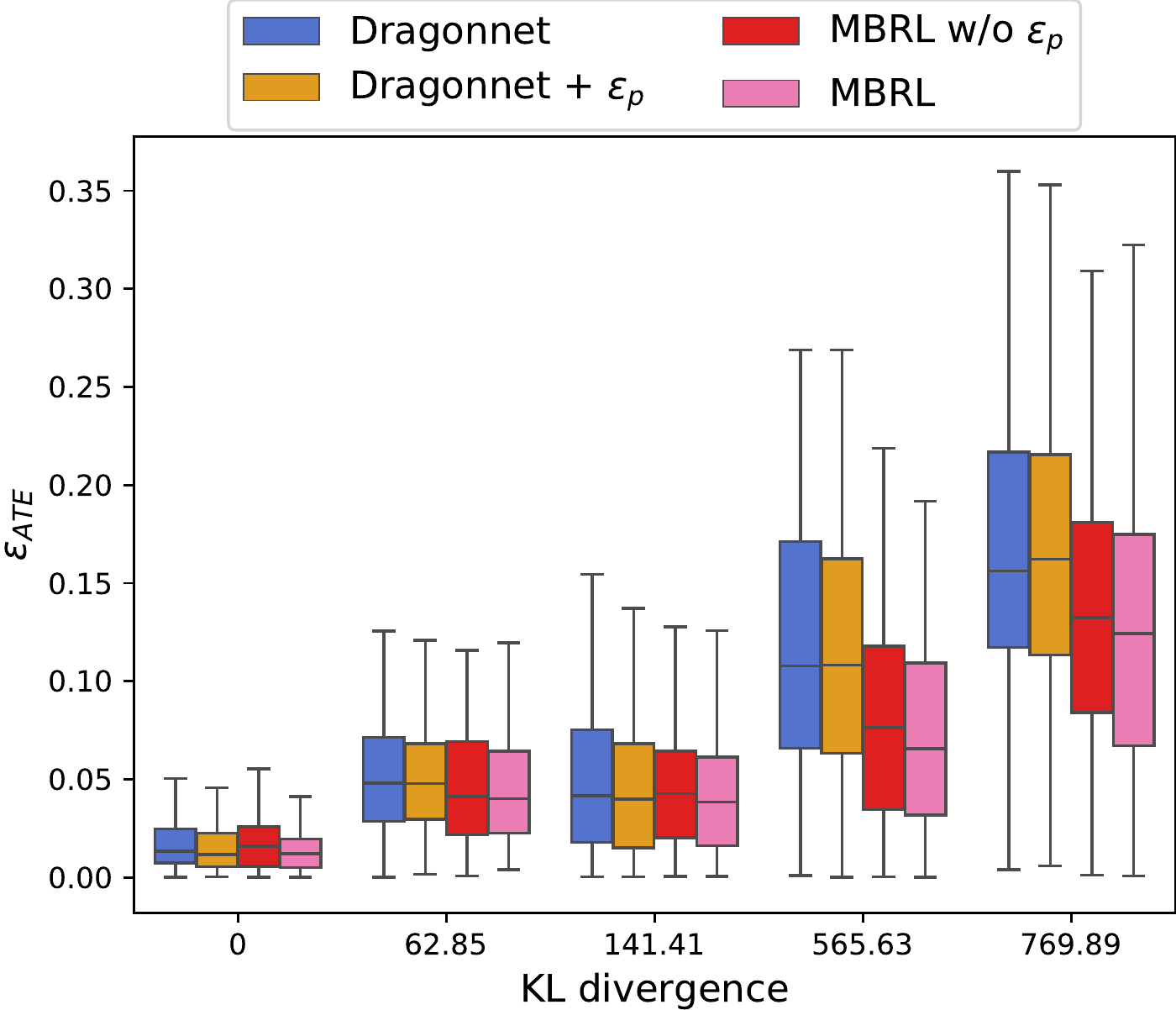}}
	\subfigure[MBRL vs. MBRL with $\theta_{1}$ and $\theta_{2}$ plugged in.]{\includegraphics[width=0.49\columnwidth]{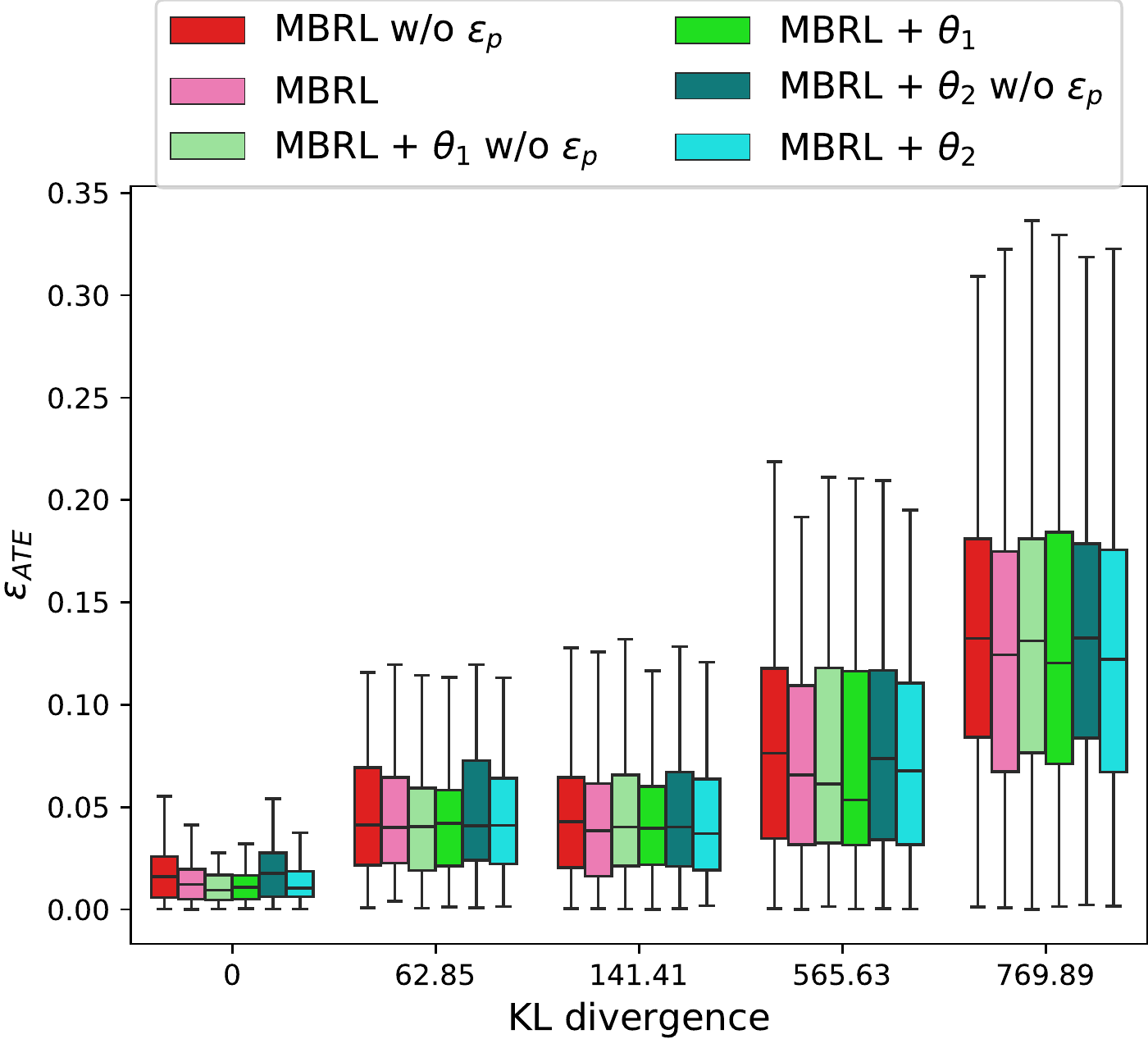}}
	\caption{Comparisons between models with and without $\epsilon_{p}$ w.r.t. varying levels of selection bias.}\label{Figure:varying_kl}
\end{figure}

We generate 2500 treated samples whose covariates $\mathbf{Z}^{1} \sim \pazocal{N}(\bm{\mu}^1, 0.5 \times \Sigma \Sigma^T)$, and 5000 controlled whose covariates $\mathbf{Z}^{0} \sim \pazocal{N}(\bm{\mu}^0, 0.5 \times \Sigma \Sigma^T)$, where $\bm{\mu}^1$ and $\bm{\mu}^0$ are both 10-dimensional vector and $\Sigma \sim \pazocal{U}((-1,1)^{10 \times 10})$. The level of selection bias, measured by KL divergence of $\bm{\mu}^1$ with respect to $\bm{\mu}^0$, would vary by fixing $\bm{\mu}^0$ and adjusting $\bm{\mu}^1$. The potential outcomes of $m^{th}$ individual are generated as $Y(1) \mid \mathbf{Z}_m \sim (\mathbf{w}_{1}^{T}\mathbf{Z}_m+n_1)$, $Y(0) \mid \mathbf{Z}_m \sim (\mathbf{w}_{0}^{T}\mathbf{Z}_m+n_0)$, where $\mathbf{w}_{1} \sim \pazocal{U}((-1,1)^{10 \times 1})$, $\mathbf{w}_{0} \sim \pazocal{U}((-1,1)^{10 \times 1})$, $n_{1} \sim \pazocal{N}(0, 0.1)$, $n_{0} \sim \pazocal{N}(0, 0.1)$. By adjusting $\bm{\mu}^1$ and fixing $\bm{\mu}^0$, we obtain five datasets with different levels of KL divergence in $\{0, \ 62.85, \ 141.41, \ 565.63, \ 769.89\}$. We run experiments on each dataset 100 times and draw box plots with regard to $\epsilon_{ATE}$ on the test set in Figure \ref{Figure:varying_kl}.

In Figure \ref{Figure:varying_kl}(a), we first find that MBRL shows stronger robustness and achieves significantly better ATE estimations with regard to different selection bias levels compared with Dragonnet. In addition, it is noticable that choosing the perturbation error $\epsilon_{p}$ as the model selection metric would yield smaller $\epsilon_{ATE}$ for any model (Dragonnet or MBRL). Particularly, $\epsilon_{p}$ corrects more errors for MBRL than Dragonnet, which indicates that $\epsilon_{p}$ works better if a model utilizes the orthogonality information in the training stage. In Figure \ref{Figure:varying_kl}(b), we have two main observations: i) the criterion $\epsilon_{p}$ improves ATE estimations for all estimators across different selection bias levels; ii) the improvement brought by $\epsilon_{p}$ becomes more substantial when selection bias increases.

\section{Related Work}
\paragraph{Representation Learning.}
Our work has a strong connection with the balanced representation learning methods proposed in \cite{johansson2016learning,shalit2017estimating}, where they mainly focus on minimizing the imbalance between the different treatment groups in the representation space but overlook maximizing the discrimination of each unit's treatment domain. IGNITE framework is proposed in \cite{guo2020ignite} to infer individual treatment effects from networked data, where they achieve a balanced representation that captures patterns of hidden confounders predictive of treatment assignments. This inspires us to study treatment effects by training a moderately-balanced representation via multi-task learning. Other works relevant to representation learning include \cite{li2017matching,louizos2017causal,yao2018representation,yoon2018ganite,shi2019adapting} and references therein.
\paragraph{Orthogonal Score Function.}
\cite{chernozhukov2018double} develop the theory of double/debiased machine learning (DML) from \cite{neyman1979c}. They define the notion of orthogonal condition, which allows their DML estimator to be doubly robust. Based on the theory of \cite{chernozhukov2018double}, another orthogonal estimator is proposed by \cite{huang2021higher}, aiming to overcome the high variance issue suffered by DML due to the misspecified propensity score. Despite the success of orthogonal estimators, the establishment of them requires the noise conditions to guarantee the corresponding score functions satisfying the orthogonal condition. None of the existing literature emphasizes the critical role of noise conditions or utilizes the orthogonality information for the model selection.
\section{Conclusion}
This paper proposes an effective representation learning method, MBRL, to study the treatment effects. Specifically, MBRL avoids the over-balanced issue by leveraging treatment domains of the representations via multi-task learning. MBRL further takes advantage of the orthogonality information and involves it in the training and validation stages. The extensive experiments show that 1) MBRL has strong predictability for the potential outcomes, distinguishability for the treatment assignment, applicability to orthogonal estimators, and robustness to the selection bias; 2) MRBL achieves substantial improvements on treatment effect estimations compared with existing state-of-the-art methods.
\subsubsection{Acknowledgements} Qi Wu acknowledges the support from the Hong Kong Research Grants Council [General Research Fund 14206117, 11219420, and 11200219], CityU SRG-Fd fund 7005300, and the support from the CityU-JD Digits Laboratory in Financial Technology and Engineering, HK Institute of Data Science. The work described in this paper was partially supported by the InnoHK initiative, The Government of the HKSAR, and the Laboratory for AI-Powered Financial Technologies.

\appendix
\section{Experiments}
The final full results for 1000 IHDP and 100 Twins experiments are reported in Table \ref{table:final result}. The mark * indicates that the baseline models did not report relevant results.

As a supplementary note, the outcome takes a binary value for Twins experiments. So the factual outcome loss in Eqn \eqref{eqt:L_FO} will be
\begin{equation*}
	{\small	\begin{aligned}
			\pazocal{L}_{fo}=-\frac{1}{N}\sum_{m=1}^{N}[y_m \log f(\Phi(\mathbf{z}_m)) + (1-y_m) \log (1-f(\Phi(\mathbf{z}_m)))].
	\end{aligned}}
\end{equation*}
\begin{table*}[htbp]
	\centering
	\caption{The full results of 1000 IHDP and 100 Twins experiments. }
	\resizebox{1\columnwidth}{!}{
		\begin{tabular}{ccccccccc}
			\toprule
			\multirow{2}[2]{*}{Method} & \multicolumn{2}{c}{IHDP In-sample} & \multicolumn{2}{c}{IHDP Out-of-sample} & \multicolumn{2}{c}{Twins In-sample} & \multicolumn{2}{c}{Twins Out-of-sample} \\
			& $\sqrt{\epsilon_{PEHE}}$ & $\epsilon_{ATE}$ & $\sqrt{\epsilon_{PEHE}}$ & $\epsilon_{ATE}$ & AUC   & $\epsilon_{ATE}$ & AUC   & $\epsilon_{ATE}$ \\
			\midrule
			OLS/LR$_1$ & $5.8 \pm .3$ & $.73 \pm .04$  & $5.8 \pm .3$ & $.94 \pm .06$ & $.660 \pm .005$ & $.004 \pm .003$ & $.500 \pm .028$ & $.007 \pm .006$ \\
			OLS/LR$_2$ & $2.4 \pm .1$ & $.14 \pm .01$ & $2.5 \pm .1$ & $.31 \pm .02$ & $.660 \pm .004$ & $.004 \pm .003$ & $.500 \pm .016$ & $.007 \pm .006$ \\
			k-NN  & $2.1 \pm .1$ & $.14 \pm .01$ & $4.1 \pm .2$ & $.79 \pm .05$ & $.609 \pm .010$ & \boldmath{}\textbf{$.003 \pm .002$}\unboldmath{} & $.492 \pm .012$ & \boldmath{}\textbf{$.005 \pm .004$}\unboldmath{} \\
			BART  & $2.1 \pm .1$ & $.23 \pm .01$ & $2.3 \pm .1$ & $.34 \pm .02$ & $.506 \pm .014$ & $.121 \pm .024$ & $.500 \pm .011$ & $.127 \pm .024$ \\
			CF    & $3.8 \pm .2$ & $.18 \pm .01$ & $3.8 \pm .2$ & $.40 \pm .03$ & *     & $.029 \pm .004$ & *     & $.034 \pm .008$ \\
			CEVAE & $2.7 \pm .1$ & $.34 \pm .01$ & $2.6 \pm .1$ & $.46 \pm .02$ & $.845 \pm .003$ & $.022 \pm .002$ & $.841 \pm .004$ & $.032 \pm .003$ \\
			SITE  & $.69 \pm .0$ & $.22 \pm .01$ & $.75 \pm .0$ & $.24 \pm .01$ & $.862 \pm .002$ & $.016 \pm .001$ & $.853 \pm .006$ & $.020 \pm .002$ \\
			GANITE & $1.9 \pm .4$ & $.43 \pm .05$ & $2.4 \pm .4$ & $.49 \pm .05$ & *     & $.006 \pm .002$ & *     & $.009 \pm .008$ \\
			BLR   & $5.8 \pm .3$ & $.72 \pm .04$ & $5.8 \pm .3$ & $.93 \pm .05$ & $.611 \pm .009$ & $.006 \pm .004$ & $.510 \pm .018$ & $.033 \pm .009$ \\
			BNN   & $2.2 \pm .1$ & $.37 \pm .03$ & $2.1 \pm .1$ & $.42 \pm .03$ & $.690 \pm .008$ & $.006 \pm .003$ & $.676 \pm .008$ & $.020 \pm .007$ \\
			TARNet & $.88 \pm .0$ & $.26 \pm .01$ & $.95 \pm .0$ & $.28 \pm .01$ & $.849 \pm .002$ & $.011 \pm .002$ & $.840 \pm .006$ & $.015 \pm .002$ \\
			CFR-WASS & $.71 \pm .0$ & $.25 \pm .01$ & $.76 \pm .0$ & $.27 \pm .01$ & $.850 \pm .002$ & $.011 \pm .002$ & $.842 \pm .005$ & $.028 \pm .003$ \\
			Dragonnet & $1.3 \pm .4$ & $.14 \pm .01$ & $ 1.3 \pm .5$ & $.20 \pm .05$ & *     & $.006 \pm .005$ & *     & $.006 \pm .005$ \\
			\midrule
			MBRL  & \boldmath{}\textbf{$.522 \pm .007$}\unboldmath{} & $.121 \pm .005$ & \boldmath{}\textbf{$.565 \pm .008$}\unboldmath{} & \boldmath{}\textbf{$.133 \pm .005$}\unboldmath{} & \boldmath{}\textbf{$.879 \pm .000$}\unboldmath{} & \boldmath{}\textbf{$.003 \pm .000$}\unboldmath{} & \boldmath{}\textbf{$.874 \pm .001$}\unboldmath{} & $.007 \pm .001$ \\
			MBRL+$\theta^i_{1}$ & \boldmath{}\textbf{$.522 \pm .007$}\unboldmath{} & \boldmath{}\textbf{$.102 \pm .004$}\unboldmath{} & \boldmath{}\textbf{$.565 \pm .008$}\unboldmath{} & $.166 \pm .007$ & \boldmath{}\textbf{$.879 \pm .000$}\unboldmath{} & \boldmath{}\textbf{$.003 \pm .000$}\unboldmath{} & \boldmath{}\textbf{$.874 \pm .001$}\unboldmath{} & $.008 \pm .000$ \\
			MBRL+$\theta^i_{2}$ & \boldmath{}\textbf{$.522 \pm .007$}\unboldmath{} & $.114 \pm .005$ & \boldmath{}\textbf{$.565 \pm .008$}\unboldmath{} & $.204 \pm .008$ & \boldmath{}\textbf{$.879 \pm .000$}\unboldmath{} & \boldmath{}\textbf{$.003 \pm .000$}\unboldmath{} & \boldmath{}\textbf{$.874 \pm .001$}\unboldmath{} & $.006 \pm .001$ \\
			\bottomrule
		\end{tabular}\label{table:final result}
	}
\end{table*}
\section{Assumptions}
\begin{assumption}[SUTVA]
	The potential outcomes for any individual are not affected by the treatment assignment of other individuals.
\end{assumption}
\begin{assumption}[Strong Ignorability]
	Given the covariates $\mathbf{Z}$, the potential outcomes are independent of the treatment assignment $D$: $\left(Y(0),Y(1)\right) \perp\!\!\!\perp D \; \mid \; \mathbf{Z}$.
\end{assumption}
\begin{assumption}[Overlap]
	The probability of treatment assignment for any unit is positive: $0<Pr(D=d \mid \mathbf{Z}=\mathbf{z})<1,$ $\forall$ $d \in \{0,1\}$ and $\mathbf{z} \in \pazocal{Z}$.
\end{assumption}
\begin{assumption}[Consistency]
	The potential outcome for treatment $d$ of each unit is equal to the observed factual outcome if the actual treatment is $d$: $(Y(d)=Y^F) \mid D=d,$ $\forall$ $d \in \{0,1\}$.
\end{assumption}

\section{Proofs}
We skip the proofs of Proposition \ref{score function} since it can be seen in \cite{chernozhukov2018double}.
In the following, we prove Proposition \ref{noise conditions} and Property \ref{noise orthogonality}.
\subsection{Proof of Proposition \ref{noise conditions}}
\begin{proof}
	The score functions stated in the Eqn. \eqref{eqt:ortho_score func_1} and Eqn. \eqref{eqt:ortho_score func_2} in the main paper are
	\begin{equation*}
		\begin{aligned}
			&\psi_{1}(W, \theta^{i}, \rho) = \theta^{i}-g(i,\mathbf{Z}) 
			-(Y-g(i,\mathbf{Z}))\frac{iD+(1-i)(1-D)}{im(\mathbf{Z})+(1-i)(1-m(\mathbf{Z}))};\\
			&\psi_{2}(W, \theta^{i}, \rho) = \theta^{i}-g(i,\mathbf{Z}) 
			-(Y(i)-g(i,\mathbf{Z}))\frac{\left((D-m(\mathbf{Z}))-\mathbb{E}\left[\nu \mid \mathbf{Z} \right]\right)^2}{\mathbb{E}\left[\nu^2 \mid \mathbf{Z}\right]}.
		\end{aligned}
	\end{equation*}
	We then check if the orthogonal condition (Definition \ref{ortho_cond}) holds for $\psi_{1}(W, \theta^{i}, \rho)$.
	\begin{equation*}
		\begin{aligned}
			&\partial_g\psi_1(W, \theta^{i}, \rho)=-1+\frac{iD+(1-i)(1-D)}{im(\mathbf{Z})+(1-i)(1-m(\mathbf{Z}))};\\
			&\partial_m\psi_1(W, \theta^{i}, \rho)=(Y-g(i,\mathbf{Z}))\frac{D}{m(\mathbf{Z})^2}, \; if \; i=1;\\
			&\partial_m\psi_1(W, \theta^{i}, \rho)=-(Y-g(i,\mathbf{Z}))\frac{1-D}{(1-m(\mathbf{Z}))^2}, \; if \; i=0.
		\end{aligned}
	\end{equation*}
	If $i=1$ and under the noise conditions $\mathbb{E}\left[\nu \mid \mathbf{Z} \right]=0$ and $\mathbb{E}\left[\xi \mid D, \mathbf{Z} \right]=0$, then we have
	\begin{equation*}
		\begin{aligned}
			&\mathbb{E}\left[\partial_g\psi_1(W, \theta^{1}, \rho) \mid \mathbf{Z} \right] \mid_{(g,m)=(g_0,m_0), \theta^{1}=\theta^{1}_{0}}\\
			&=-1+\mathbb{E}\left[\frac{D}{m_0(\mathbf{Z})} \mid \mathbf{Z} \right]\\
			&=-1+\mathbb{E}\left[\frac{D-m_0(\mathbf{Z})+m_0(\mathbf{Z})}{m_0(\mathbf{Z})} \mid \mathbf{Z} \right]\\
			&=-1+\mathbb{E}\left[\frac{\nu+m_0(\mathbf{Z})}{m_0(\mathbf{Z})} \mid \mathbf{Z} \right]\\
			&=-1+\frac{\mathbb{E}\left[\nu \mid \mathbf{Z}\right]}{m_0(\mathbf{Z})} + 1\\
			&=0.
		\end{aligned}
	\end{equation*}
	\begin{equation*}
		\begin{aligned}
			&\mathbb{E}\left[\partial_m\psi_1(W, \theta^{1}, \rho) \mid \mathbf{Z} \right] \mid_{(g,m)=(g_0,m_0), \theta^{1}=\theta^{1}_{0}}\\
			&=\mathbb{E}\left[(Y-g_0(i,\mathbf{Z}))\frac{D}{m_0(\mathbf{Z})^2} \mid \mathbf{Z}\right]\\
			&=\mathbb{E}\left[\mathbb{E}\left[(Y-g_0(i,\mathbf{Z}))\frac{D}{m_0(\mathbf{Z})^2} \mid D, \mathbf{Z}\right] \mid \mathbf{Z}\right]\\
			&=\mathbb{E}\left[\mathbb{E}\left[\xi\frac{D}{m_0(\mathbf{Z})^2} \mid D, \mathbf{Z}\right] \mid \mathbf{Z}\right]\\
			&=\mathbb{E}\left[\mathbb{E}\left[\xi\mid D, \mathbf{Z}\right]\frac{D}{m_0(\mathbf{Z})^2} \mid \mathbf{Z}\right]\\
			&=0.
		\end{aligned}
	\end{equation*}
	If $i=0$ and under the noise conditions $\mathbb{E}\left[\nu \mid \mathbf{Z} \right]=0$ and $\mathbb{E}\left[\xi \mid D, \mathbf{Z} \right]=0$, then we have
	\begin{equation*}
		\begin{aligned}
			&\mathbb{E}\left[\partial_g\psi_1(W, \theta^{0}, \rho) \mid \mathbf{Z} \right] \mid_{(g,m)=(g_0,m_0), \theta^{0}=\theta^{0}_{0}}\\
			&=-1+\mathbb{E}\left[\frac{1-D}{1-m_0(\mathbf{Z})} \mid \mathbf{Z} \right]\\
			&=-1+\mathbb{E}\left[\frac{1-D-m_0(\mathbf{Z})+m_0(\mathbf{Z})}{1-m_0(\mathbf{Z})} \mid \mathbf{Z} \right]\\
			&=-1+\mathbb{E}\left[\frac{-\nu+1-m_0(\mathbf{Z})}{1-m_0(\mathbf{Z})} \mid \mathbf{Z} \right]\\
			&=-1-\frac{\mathbb{E}\left[\nu \mid \mathbf{Z}\right]}{1-m_0(\mathbf{Z})} + 1\\
			&=0.
		\end{aligned}
	\end{equation*}
	\begin{equation*}
		\begin{aligned}
			&\mathbb{E}\left[\partial_m\psi_1(W, \theta^{0}, \rho) \mid \mathbf{Z} \right] \mid_{(g,m)=(g_0,m_0), \theta^{0}=\theta^{0}_{0}}\\
			&=\mathbb{E}\left[-(Y-g_0(i,\mathbf{Z}))\frac{1-D}{(1-m_0(\mathbf{Z}))^2} \mid \mathbf{Z}\right]\\
			&=\mathbb{E}\left[\mathbb{E}\left[-(Y-g_0(i,\mathbf{Z}))\frac{1-D}{(1-m_0(\mathbf{Z}))^2} \mid D, \mathbf{Z}\right] \mid \mathbf{Z}\right]\\
			&=\mathbb{E}\left[\mathbb{E}\left[-\xi\frac{1-D}{(1-m_0(\mathbf{Z}))^2} \mid D, \mathbf{Z}\right] \mid \mathbf{Z}\right]\\
			&=\mathbb{E}\left[-\mathbb{E}\left[\xi\mid D, \mathbf{Z}\right]\frac{1-D}{(1-m_0(\mathbf{Z}))^2} \mid \mathbf{Z}\right]\\
			&=0.
		\end{aligned}
	\end{equation*}
	We then check if the orthogonal condition (Definition \ref{ortho_cond}) holds for $\psi_{2}(W, \theta^{i}, \rho)$.
	\begin{equation*}
		\begin{aligned}
				\partial_m\psi_2(W, \theta^{i}, \rho)=(Y(i)-g(i,\mathbf{Z}))\frac{2((D-m(\mathbf{Z}))-\mathbb{E}\left[\nu \mid \mathbf{Z} \right])}{\mathbb{E}\left[\nu^2 \mid \mathbf{Z}\right]}
		\end{aligned}
	\end{equation*}
	\begin{equation*}
		\begin{aligned}
			\partial_g\psi_2(W, \theta^{i}, \rho)=-1+\frac{\left((D-m(\mathbf{Z}))-\mathbb{E}\left[\nu \mid \mathbf{Z} \right]\right)^2}{\mathbb{E}\left[\nu^2 \mid \mathbf{Z}\right]}
		\end{aligned}
	\end{equation*}
	Using the noise condition $\mathbb{E}\left[\nu \mid \mathbf{Z} \right]=0$, we have
	\begin{equation*}
		\begin{aligned}
			&\mathbb{E}\left[\partial_g\psi_2(W, \theta^{i}, \rho) \mid \mathbf{Z} \right] \mid_{(g,m)=(g_0,m_0), \theta^{i}=\theta^{i}_{0}}\\
			&=-1+\mathbb{E}\left[\frac{\left((D-m_0(\mathbf{Z}))-\mathbb{E}\left[\nu \mid \mathbf{Z} \right]\right)^2}{\mathbb{E}\left[\nu^2 \mid \mathbf{Z}\right]} \mid \mathbf{Z} \right]\\
			&=-1 + \frac{1}{\mathbb{E}\left[\nu^2 \mid \mathbf{Z}\right]}\mathbb{E}\left[\left((D-m_0(\mathbf{Z}))-\mathbb{E}\left[\nu \mid \mathbf{Z} \right]\right)^2 \mid \mathbf{Z} \right]\\
			&=-1 + \frac{1}{\mathbb{E}\left[\nu^2 \mid \mathbf{Z}\right]}\mathbb{E}[ (D-m_0(\mathbf{Z}))^2 + (\mathbb{E}\left[\nu \mid \mathbf{Z} \right])^2
			- 2(D-m_0(\mathbf{Z}))\mathbb{E}\left[\nu \mid \mathbf{Z} \right] \mid \mathbf{Z}]\\
			&=-1 + \frac{1}{\mathbb{E}\left[\nu^2 \mid \mathbf{Z}\right]}\left[\mathbb{E}\left[\nu^2 \mid \mathbf{Z}\right] + (\mathbb{E}\left[\nu \mid \mathbf{Z} \right])^2 - 2(\mathbb{E}\left[\nu \mid \mathbf{Z} \right])^2\right]\\
			&=-1 + \frac{1}{\mathbb{E}\left[\nu^2 \mid \mathbf{Z}\right]}\left[\mathbb{E}\left[\nu^2 \mid \mathbf{Z}\right] - (\mathbb{E}\left[\nu \mid \mathbf{Z} \right])^2\right]\\
			&=-1 + \frac{\mathbb{E}\left[\nu^2 \mid \mathbf{Z}\right]}{\mathbb{E}\left[\nu^2 \mid \mathbf{Z}\right]}=0.
		\end{aligned}
	\end{equation*}
	By the model setup $Y=g_0(D, \mathbf{Z}) + \xi$, we have the underlying relation for the potential outcome $Y(i)$ that $Y(i)=g_0(i, \mathbf{Z}) + \xi$. Using the noise condition $\mathbb{E}\left[\xi \mid D, \mathbf{Z} \right]=0$, we have
	\begin{equation*}
		\begin{aligned}
			&\mathbb{E}\left[\partial_m\psi_2(W, \theta^{i}, \rho) \mid \mathbf{Z} \right]\mid_{(g,m)=(g_0,m_0), \theta^{i}=\theta^{i}_{0}}\\
			&=\mathbb{E}\left[\mathbb{E}\left[\xi\frac{2((D-m_0(\mathbf{Z}))-\mathbb{E}\left[\nu \mid \mathbf{Z} \right])}{\mathbb{E}\left[\nu^2 \mid \mathbf{Z}\right]} \mid D, \mathbf{Z} \right] \mid \mathbf{Z} \right]\\
			&=\mathbb{E}\left[\frac{2((D-m_0(\mathbf{Z}))-\mathbb{E}\left[\nu \mid \mathbf{Z} \right])}{\mathbb{E}\left[\nu^2 \mid \mathbf{Z}\right]}\mathbb{E}\left[\xi \mid D, \mathbf{Z} \right] \mid \mathbf{Z} \right]\\
			&=0.
		\end{aligned}
	\end{equation*}
	Therefore, the noise conditions $\mathbb{E}\left[\xi \mid D, \mathbf{Z} \right]=0$ and $\mathbb{E}\left[\nu \mid \mathbf{Z} \right]=0$ are sufficient for the score functions $\psi_1$ and $\psi_2$ satisfying the orthogonal condition.
\end{proof}
\subsection{Proof of Property \ref{noise orthogonality}}
\begin{proof}
	Using the noise condition $\mathbb{E}\left[\xi \mid D, \mathbf{Z} \right]=0$, we have
	\begin{equation*}
		\begin{aligned}
			&\mathbb{E}\left[(Y-g_0(D, \mathbf{Z}))(D-m_0(\mathbf{Z}))\right]\\
			=&\mathbb{E}\left[\mathbb{E}\left[(Y-g_0(D, \mathbf{Z}))(D-m_0(\mathbf{Z})) \mid D, \mathbf{Z}\right]\right]\\
			=&\mathbb{E}\left[(D-m_0(\mathbf{Z}))\mathbb{E}\left[(Y-g_0(D, \mathbf{Z})) \mid D, \mathbf{Z}\right]\right]\\
			=&\mathbb{E}\left[(D-m_0(\mathbf{Z}))\mathbb{E}\left[\xi \mid D, \mathbf{Z}\right]\right]\\
			=&0.
		\end{aligned}
	\end{equation*}
\end{proof}
\bibliographystyle{splncs04}
\bibliography{pricai2022}
\end{document}